\documentclass[runningheads]{llncs}\usepackage{graphicx}
\usepackage{algorithm}  
\usepackage{algorithmicx}  
\usepackage{algpseudocode}  
\usepackage{amsmath} 
\usepackage{amsfonts}
\usepackage{booktabs}
\usepackage{multirow}
\usepackage{regexpatch}
\usepackage{amssymb}
\usepackage{mathrsfs}
\usepackage[misc]{ifsym} 
\usepackage[marginal]{footmisc}

\usepackage{graphicx}
\usepackage{float} 
\usepackage{subfigure}

\begin{document}

\title{Recurrence-free Survival Prediction under the Guidance of Automatic Gross Tumor Volume Segmentation for Head and Neck Cancers}
\titlerunning{HNC GTV Segmentation Guided RFS Prediction}
\author{Kai Wang\orcidID{0000-0003-2155-1445}, Yunxiang Li\orcidID{0000-0003-0622-4710}, Michael Dohopolski\orcidID{0000-0002-9043-1490},Tao Peng\orcidID{0000-0003-0848-7901}, Weiguo Lu\orcidID{0000-0003-3036-7287}, You Zhang\orcidID{0000-0002-8033-2755}, and Jing Wang\orcidID{0000-0002-8491-4146}\Letter}
\authorrunning{K. Wang et al.}
\institute{Department of Radiation Oncology, University of Texas Southwestern Medical Center, Dallas, USA \\  \email{Jing.Wang@UTSouthwestern.edu} \\
\url{https://labs.utsouthwestern.edu/advanced-imaging-and-informatics-radiation-therapy-airt-lab}}
\maketitle

\begin{abstract}
For Head and Neck Cancers (HNC) patient management, automatic gross tumor volume (GTV) segmentation and accurate pre-treatment cancer recurrence prediction are of great importance to assist physicians in designing personalized management plans, which have the potential to improve the treatment outcome and quality of life for HNC patients. In this paper, we developed an automated primary tumor (GTVp) and lymph nodes (GTVn) segmentation method based on combined pre-treatment positron emission tomography/computed tomography (PET/CT) scans of HNC patients. We extracted radiomics features from the segmented tumor volume and constructed a multi-modality tumor recurrence-free survival (RFS) prediction model, which fused the prediction results from separate CT radiomics, PET radiomics, and clinical models. We performed 5-fold cross-validation to train and evaluate our methods on the MICCAI 2022 HEad and neCK TumOR segmentation and outcome prediction challenge (HECKTOR) dataset. The ensemble prediction results on the testing cohort achieved Dice scores of 0.77 and 0.73 for GTVp and GTVn segmentation, respectively, and a C-index value of 0.67 for RFS prediction. The code is publicly available (\url{https://github.com/wangkaiwan/HECKTOR-2022-AIRT}). Our team's name is AIRT.

\end{abstract}
\footnote{Kai Wang and Yunxiang Li—Equal contribution.}

\section{Introduction}\label{sec:intro}

\begin{figure}[ht]
  \centering
  \includegraphics[width=0.9\textwidth]{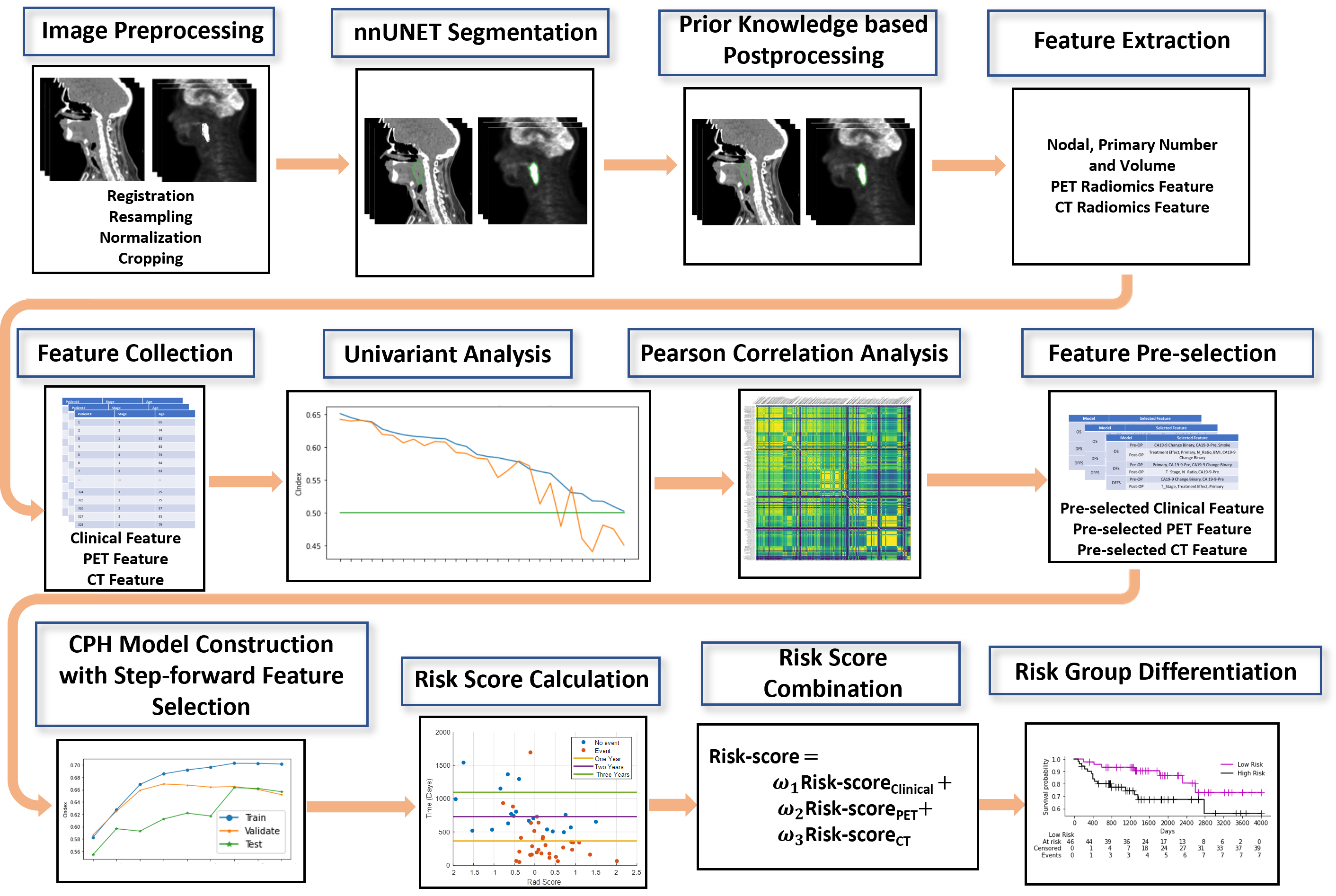}
  \caption{A schematic overview of our proposed automatic gross tumor volume segmentation guided cancer recurrence free survival prediction framework.}
  \label{fig:workflow}
\end{figure}

Head and Neck cancer (HNC) is one of the most common cancers worldwide \cite{jemal2011global}. With the development of HNC radiotherapy and chemo-radiotherapy, HNC patients can be successfully treated in many cases. However, even treated with curative intent, there are still more than 15\% patients will experience cancer recurrence, of which locoregional failures occur in up to 40\% of patients in the first two years after the treatment \cite{denaro2016follow}. Therefore, a strategy that can accurately identify HNC patients at high risk of recurrence at diagnosis would be helpful in assisting physicians in making personalized treatment plans, which have the potential to improve patient treatment outcomes.

Positron Emission Tomography (PET) and Computed Tomography (CT) imaging (PET/CT) play an important role in the management of HNC patients \cite{goel2017clinical}. At initial staging, PET/CT plays the central role in characterizing local, regional and distant disease, while during therapy and after therapy, it is routinely used to assess the treatment response and detect recurrence and/or metastases. Recently, using PET/CT scans of HNC patients, several radiomics and deep learning studies were proposed to better predict the treatment outcomes, including locoregional recurrence, distant recurrence, and recurrence-free survival (RFS) \cite{vallieres2017radiomics,diamant2019deep,wang2020multi,chen2021attention,wang2022locoregional}. Besides, as deep learning-based image segmentation has achieved comparable performance to humans in many tasks, several organs at risk (OAR) and gross tumor volume (GTV) automatic segmentation methods were proposed to help in the management of HNC patients \cite{van2019deep,zhu2019anatomynet,nikolov2018deep}. These segmentation methods not only have the potential to ease the workload of healthcare workers but also assist in the workflow of radiomics or deep learning-based treatment outcome prediction models, which can heavily rely on the accuracy of the region of interest (ROI) delineation.

Although the promising performance was reported in some of the automatic segmentation and outcome prediction studies, their reliability, robustness, and reproducibility are still of concern for clinical translation. Aiming to further study the automatic segmentation and RFS survival prediction method for HNC tumors and exam their cross-institutional performance objectively, the MICCAI 2022 Head and NeCK TumOR segmentation and outcome prediction challenge (HECKTOR) offered researchers the opportunity to build and evaluate their auto-segmentation and RFS prediction models with a very large dataset collected from multiple institutions \cite{oreiller2022head,andrearczyk2022overview}. In the present work, we built an auto-segmentation framework for precise segmentation of HNC primary tumor (GTVp) and lymph nodes (GTVn) with HECKTOR 2022 dataset, and we introduced a multi-modality HNC RFS prediction framework that ensembles the predicted risk scores from separate clinical feature model, PET radiomics model, and CT radiomics model. The deep auto-segmentation model is used to identify the ROIs for radiomics feature extraction.

\section{Material and Methods}
\subsection{Head and NeCK TumOR 2022 (HECKTOR 2022) Dataset}
The dataset provided in HECKTOR 2022 consisted of PET/CT scans, clinical features, and follow-up information of 845 HNC patients from 7 centers. The total number of training cohorts is 524 and 489 for tasks 1 (GTVp and GTVn segmentation) and 2 (RFS prediction), respectively. Training and test cohorts are representative of the distribution of the real-world population of patients accepted for initial staging of oropharyngeal cancer (with around 21\% of recurrence and a median RFS of 14 months in the training set). For the PET images, computation of the Standardized Uptake Value (SUV) is already done by the challenge organizer. Missing patient weight was estimated as 75Kg to compute SUV for a small subset of patients (8 of 845). All the PET/CT files conversed from the DICOM file format to NIfTI format. More detailed description of the dataset can be found \cite{oreiller2022head,andrearczyk2022overview}. The provided clinical features are gender, age, tobacco and alcohol consumption, performance status (Zubrod), HPV status, treatment (surgery and/or chemotherapy in addition to the radiotherapy that all patients underwent). There is some clinical information missing (tobacco, alcohol, performance status, HPV status), and we encoded the missing feature with value 0, known as negative status as -1. Five-fold cross-validation strategy was used to train and validate our method for both two tasks. And we fused the prediction results of 5-fold models for the testing cohort to generate the ensemble result for the challenge submission. 

\subsection{Overall Architecture}
The workflow of our proposed method is shown in Fig. \ref{fig:workflow}. After PET/CT image processing, we trained a nnUNET model with concatenated PET/CT image as the input, and the output is GTVp and GTVn masks. Then we calculated the predicted GTVp and GTVn number and volume as additional clinical features. Within the predicted GTVp volume, we extracted PET, and CT radiomics feature separately. After 
univariate analysis and feature correlation analysis, we removed the low-predictive features and redundant features. The remaining features would go through a step-forward feature selection with Cox Proportional Hazards (CPH) model, and the best model and feature set was determined by the validation performance. Then the risk score of testing patients can be calculated. We trained the clinical feature model, CT radiomics model, and PET radiomics model separately. Their predicted risk score for testing data was fused together as the multi-modality RFS model prediction results to predict patients' RFS and identify high- or low-risk patients.

\subsection{Gross Tumor Volume Segmentation (Task 1)}
\subsubsection{nnUNet for Segmentation}
We used the nnUNet \cite{isensee2021nnu} 3D full-resolution pipeline as the basis segmentation backbone. Considering each patient image had a different voxel size, the original PET/CT images were resampled to 2 mm × 2 mm × 2 mm pixel spacing with trilinear interpolation. The labels were resampled to the same size spacing.
Besides, the PET/CT images were cropped into 2 × 128 × 128 × 128 patches with overlaps. Each image was standardized using z-scores. To avoid overfitting, image augmentation was used with random rotation in all directions between [-30, 30] degrees and random scaling between  [0.7, 1.4]. All 524 patients were randomly divided into five groups for the five-fold cross-validation, each time leaving out one group of patients for validation and the others for training. 

The training protocol is listed in Table. \ref{tab:training_parameters}.

\begin{table}[ht]
\caption{Training protocols used for training and optimization of our method.}
\begin{tabular}{l|l} 
\hline
Parameter                              & Setting               \\ \hline 
Patch size                             & 128 × 128 × 128                             \\
Batch size                             & 2                                        \\
weight\_decay                            &  3e-05                                   \\
Optimizer SGD                          & SGD                                          \\
Initial learning rate 0.01             & 0.01                                    \\
Learning rate decay schedule           & polyLR                                  \\
Nesterov momentum                      & 0.99                                         \\
Epoch                                  & 1000                     \\  \hline
\end{tabular}
\centering
\label{tab:training_parameters}

\end{table}

\subsubsection{Post-processing}
It is known a priori from anatomical knowledge that the distances between the primary tumor (GTVp) and lymph nodes (GTVn) are often under certain thresholds. In order to remove unreasonable segmentation results, we first calculate the center-of-mass coordinates of GTVp and GTVn. Assume the center-of-mass coordinates of GTVp and GTVn are $(x_p,y_p,z_p)$ and $(x_n,y_n,z_n)$, separately. The distance $D_{pn}$ between GTVp and GTVn can be calculate via Eq. (\ref{distance}).

\begin{equation} \label{distance}
\begin{aligned}
    D_{pn}= \sqrt{(x_p-x_n)^2+(y_p-y_n)^2+(z_p-z_n)^2}
\end{aligned}
\end{equation}
Based on the clinician's empirical judgment, we set $D_{max}$ as the threshold value for distance. Then, we remove the lymph nodes whose distance is greater than $D_{max}$.

\subsection{Recurrence-Free Survival Prediction (Task 2)}
With the nnUNET predicted the GTVp and GTVn volume on the resampled PET/CT image, we extract PET and CT radiomics feature separately with the pyradiomics package \cite{van2017computational}. Default feature extraction settings were used for both PET and CT feature extraction. As the number and volume of GTVp and GTVn are reported to be predictive for HNC treatment outcome prediction, and they are not as complex features as radiomics features, we added them to clinical feature set. For the missing value in clinical feature, comprising 293 tobacco consumption, 326 alcohol consumption, 268 performance status, and 167 missing HPV status, we coded the positive status (1 in the original file) to 1, negative status (0 in the original file) to -1, and missing value to 0.

For training and validating our method with training cohort data, we followed the same splitting as what we did for training the nnUNET. For each training fold in the 5-fold cross-validation procedure, we first performed 100 times internal 5-fold cross-validation on the training fold for clinical feature, CT radiomcis feature, PET radiomics feature pre-selection separately. The partition of the internal training and validation is random each time. Univariant cox regression models were built with each feature for RFS prediction. The average C-index of univariant RFS prediction models validation data was recorded, and the order of features was sorted based on the validation performance. To avoid the low-predictive-ability features, the feature that has a lower than 0.5 average C-index was removed from the feature set. To reduce redundancy in the feature sets for different prediction targets, we performed Pearson correlation analysis for all the features. A feature that has an absolute correlation coefficient higher than 0.9 to any of its previous features was removed from the feature set.

Then, we performed multivariate CPH regression with the step forward feature selection strategy to further select predictive features and construct the RFS prediction models. C-index was the criteria for the step forward feature selection in this step, 5 was set as the maximum number of selected clinical features, 10 was set as the maximum number of selected radiomics features. Another internal 100 times 5-fold cross-validation was conducted to mitigate the impact of random patient partition. The risk scores of training samples and validation samples in each time of 5-fold cross-validation were recorded. The combination of the 5-fold risk scores was used as the final survival prediction on the training cohort of each single-modality model, and the C-index was used to evaluate our models. We then calculated the average risk scores of the three single-modality models for testing patients and used them as our final prediction (multi-modality RFS prediction model results). Of note, for patients who were predicted to have no GTVp volume, their final prediction results are the same as clinical feature-based models, as they don't have GTVp radiomics features. We also used the multi-modality risk score to differentiate between high- and low-risk patient groups. The threshold was set as 0. We performed Kaplan–Meier analysis and log-rank test to evaluate the RFS difference of the identified patient groups, a P-value $\leqslant 0.05$ was considered significant. Lifelines python package was used to construct the model and perform statistic analysis.

\begin{figure}[ht]
    \centering
    \includegraphics[width=0.99\textwidth]{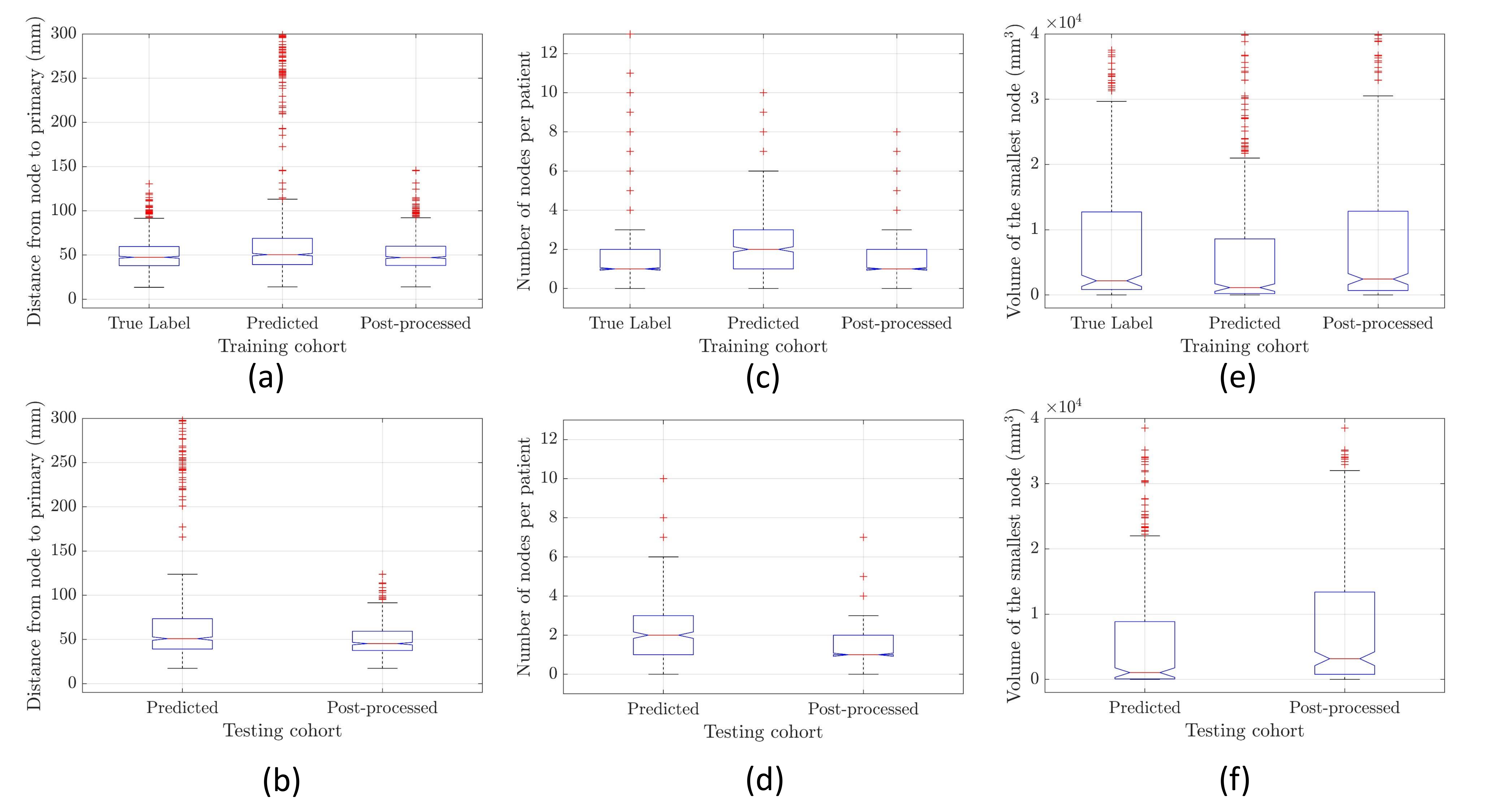}
    \caption{Node properties of label, predicted and post-processed data on training (a,c,e) and testing (b,d,f) cohorts. The analyzed properties include distance from nodes to center of primary tumor for each patients (a,b), number of nodes per patients (c,d), and volume of the smallest node of each patient (e,f).}
    \label{fig:boxplot}
\end{figure}

\section{Results}

\subsection{Gross Tumor Volume Segmentation (Task 1)} 

\begin{figure}[ht]
  \centering
  \includegraphics[width=0.99\textwidth]{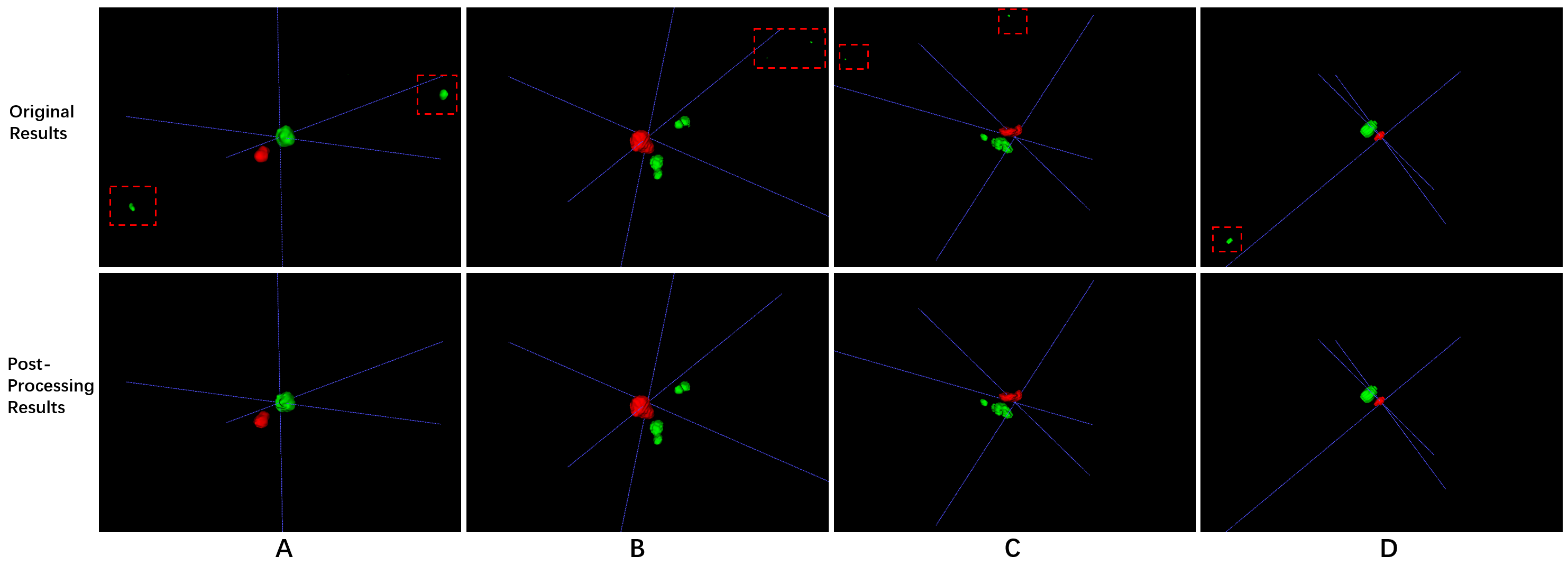}
  \caption{Visualization comparison of the post-processed results and the original results. Obvious error areas are marked with red boxes.}
  \label{fig:comparsion_post}
\end{figure}

This may be due to the fact that the data contains most of the body parts and the lymph nodes are only distributed in small locations, and other parts of the body to be easily misidentified, which is illustrated in Fig. \ref{fig:comparsion_post}. Our method is capable of identifying the segmentation result of other parts as unreliable results and excluding it.
Besides, as shown in Fig. \ref{fig:boxplot}, the data distribution of several metrics becomes closer to the manual label, including the number of nodes, volume of the smallest node, and distance from the node to the primary. 
As shown in Table. \ref{tab:seg_post}, we quantitatively compare the segmentation results before and after post-processing, and we can see that our DICE has some improvement after processing.
\begin{table}[ht]
\caption{Nodal and primary tumor segmentation Dice similarity coefficients before and after post-processing on testing cohort.}
\centering
\begin{tabular}{c|c|c|c} \hline
             & GTVn DICE & GTVp DICE & mean DICE \\ \hline
Original     & 0.73276     & 0.76689     & 0.74983                      \\ \hline
Post-process & 0.73392     & 0.76689     & 0.75040                     \\ \hline
\end{tabular}
\label{tab:seg_post}
\end{table}

\subsection{Recurrence-Free Survival Prediction (Task 2)}

\begin{figure}[ht]
    \centering
    \subfigure[Clinical Feature Model]{
    \includegraphics[width=0.45\textwidth]{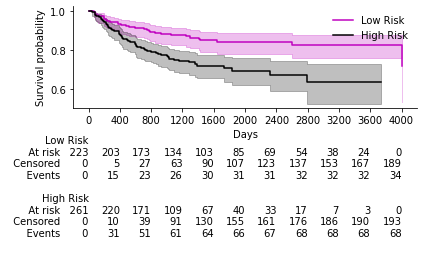}
    \centering
    }
    \subfigure[PET Radiomics Model]{
    \includegraphics[width=0.45\textwidth]{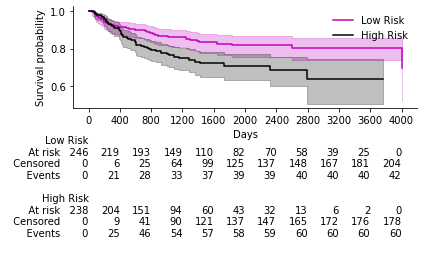}
    \centering
    }
    \subfigure[CT Radiomics Model]{
    \includegraphics[width=0.45\textwidth]{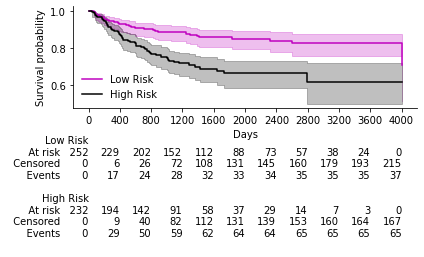}
    \centering
    }
    \subfigure[Multi-modality Model]{
    \includegraphics[width=0.45\textwidth]{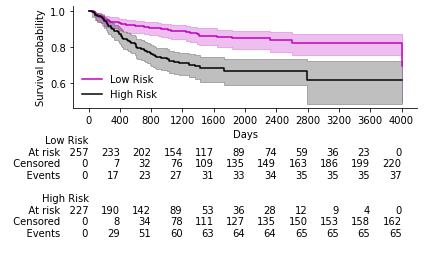}
    \centering
    }
    
    \caption{Kaplan–Meier analysis of risk scores from single- and multi-modality recurrence-free survival (RFS) models on training cohort. A risk score value of 0 was used to identify the high- and low-risk patient group. For all four models, the identified low-risk groups have significantly better RFS than the high-risk group, the $-log2(p)$ values of them are 11.84, 8.05, 15.89, and 16.98, respectively. }
    \label{fig:rfs_KM}
\end{figure}

The RFS prediction performance evaluated with C-index on training and testing cohort is shown in Fig. \ref{tab:rfs_results}. The Kaplan–Meier analysis and log-rank test results of the RFS difference of different patient risk groups identified by single-modality and multi-modality RFS models are shown in Fig. \ref{fig:rfs_KM}. To our surprise, although there are a lot of missing data in the clinical feature set, the clinical feature-based model is still very predictive (C-index=0.68 on training cohort), and the top predictive clinical features are GTVp volume, Node volume, HPV status, and tobacco consumption. The CT-radiomics model is the most predictive single-modality model, and its C-index on the training cohort is 0.69. The fused multi-modality RFS prediction model has the best prediction performance in the training cohort, with a prediction C-index value of 0.72. On the testing cohort, our submitted multi-modality RFS risk score got a C-index value of 0.67, which ranked No.3 on the challenge leaderboard. 

\begin{table}[ht]
\caption{Single-modality and Multi-modality models performance for recurrence-free survival prediction on training and testing cohort.}
\centering
\begin{tabular}{c|c|c|c|c} \hline
Modality               & Clinical    & CT Radiomics & PET Radiomics & Multi-Modality \\ \hline
C-index (Training)     & 0.68253     & 0.68927      & 0.62535       & 0.72036               \\ \hline
C-index (Testing)      & ---         & ---          & ---           & 0.67257               \\ \hline
\end{tabular}
\label{tab:rfs_results}
\end{table}

\section{Discussion and Conclusion}

In this paper, we presented a multi-modality HNC RFS prediction method guided by auto-segmentation GTV segmentation, which achieved a 0.67 C-index for RFS prediction and 0.75 DSC for GTV segmentation. Although the results got good rankings in the challenge, there are still some limitations to our work. Firstly, our segmentation-guided RFS prediction model is not optimized end-to-end, current separate training strategy might not yield the best performance. Secondly, as radiomics method heavily relies on accurate ROI delineation, it's important to evaluate the performance difference between the model built with features extracted from human expert delineation and features extracted from auto-segmented volume. Thirdly, as RFS prediction is a medical task, irresponsible prediction can be costly if a RFS prediction model is used in the clinic. However, our current RFS model can only give a one-value risk prediction without reporting any uncertainty about the corresponding prediction. It's hard to tell whether the prediction is based on learning or just random guessing \cite{https://doi.org/10.48550/arxiv.2208.14452}. Uncertainty analysis would be one of our future work, especially with a large dataset as HECKTOR 2022. 

\textbf{Acknowledgements.} This work was supported in part by National Institutes of Health (Grant No. R01CA251792, R01CA240808 and R01CA258987)

\bibliographystyle{unsrt}
\begingroup
  
  \small 
  \bibliography{paper}
\endgroup

\end{document}